\documentclass[sigconf]{acmart}
\AtBeginDocument{%
  }

\usepackage{amsmath}
\usepackage{booktabs}
\usepackage{caption}
\usepackage{subcaption}
\usepackage{multirow}
\usepackage{diagbox}
\usepackage{colortbl}
\usepackage{algorithm}
\usepackage{adjustbox}
\usepackage{algpseudocode}
\usepackage{graphicx}
\usepackage{dirtytalk}
\usepackage{comment}

\copyrightyear{2025}
\acmYear{2025}
\setcopyright{acmlicensed}
\acmConference[CIKM '25] {Proceedings of the 34th ACM International Conference on Information and Knowledge Management}{ November 10--14, 2025}{Seoul, Republic of Korea.}
\acmBooktitle{Proceedings of the 34th ACM International Conference on Information and Knowledge Management (CIKM '25), November 10--14, 2025, Seoul, Republic of Korea}
\acmISBN{979-8-4007-2040-6/2025/11}
\acmDOI{10.1145/XXXXXX.XXXXXX}
\begin{document}

\title{Multimodal RAG Enhanced Visual Description}

\author{Amit Kumar Jaiswal}
%\email{webmaster@marysville-ohio.com}
\affiliation{%
  \institution{Indian Institute of Technology (BHU)}
  \city{Varanasi}
  \country{India}
}
\email{amit.chr@iitbhu.ac.in}

\author{Haiming Liu}
\affiliation{%
  \institution{University of Southampton}
  %%\city{}
  \country{United Kingdom}}
\email{h.liu@soton.ac.uk}

\author{Ingo Frommholz}
\affiliation{%
  \institution{Modul University Vienna}
  \city{Vienna}
  \country{Austria}
}
\email{ifrommholz@acm.org}

\renewcommand{\shortauthors}{Jaiswal et al.}

\begin{abstract}
Textual descriptions for multimodal inputs entail recurrent refinement of queries to produce relevant output images. Despite efforts to address challenges such as scaling model size and data volume, the cost associated with pre-training and fine-tuning remains substantial. However, pre-trained large multimodal models (LMMs) encounter a modality gap, characterised by a misalignment between textual and visual representations within a common embedding space. Although fine-tuning can potentially mitigate this gap, it is typically expensive and impractical due to the requirement for extensive domain-driven data. To overcome this challenge, we propose a lightweight training-free approach utilising Retrieval-Augmented Generation (RAG) to extend across the modality using a linear mapping, which can be computed efficiently. During inference, this mapping is applied to images embedded by an LMM enabling retrieval of closest textual descriptions from the training set. These textual descriptions, in conjunction with an instruction, cater as an input prompt for the language model to generate new textual descriptions. In addition, we introduce an iterative technique for distilling the mapping by generating synthetic descriptions via the language model facilitating optimisation for standard utilised image description measures. Experimental results on two benchmark multimodal datasets demonstrate significant improvements.
%The creation of textual descriptions for multimodal inputs entails recurrent refinement of queries to produce relevant output images. Despite efforts to address challenges such as scaling model size and data volume, the cost associated with pre-training and fine-tuning remains substantial. However, pre-trained large multimodal models (LMMs) encounter a modality gap, characterised by a misalignment between textual and visual representations within a common embedding space. Although fine-tuning can potentially mitigate this gap, it is typically expensive and impractical due to the requirement for extensive domain-driven data. To overcome this challenge, we propose a lightweight training-free approach utilising Retrieval-Augmented Generation (RAG) to extend across the modality using a linear mapping, which can be computed efficiently. During inference, this mapping is applied to images embedded by an LMM, enabling retrieval of closest textual descriptions from the training set. These textual descriptions in conjunction with an instruction cater as an input prompt for the language model to generate new textual descriptions. In addition, we introduce an iterative technique for distilling the mapping by generating synthetic descriptions via the language model, facilitating optimisation for standard utilised image description measures. Experimental results on two benchmark datasets demonstrate the efficacy of our method.
\end{abstract}




%Textual descriptions generated via \emph{mRAG-gim} uses image collections from the MSCOCO validation set which follows our computed mapping trained through the MSCOCO training set.
\maketitle

\section{Introduction and Background}
%\vspace{-6pt}
Traditional image captioning tasks involve generating textual descriptions that align with the distribution of reference image-text pairs. Such visual descriptions serve several purposes, including image search, content-based image retrieval, and enhancing readiness for dyslexic users with visual impairments~\cite{sidorov2020textcaps}. However, the brevity of textual descriptions in typical multimodal datasets might fail to distinctly identify the associated images. This attenuates upon utilising models which are trained specifically on image-text pairs acquired from web sources. Generative approaches such as RAG are able to comprehend the interplay among textual and visual elements by necessitating generative adeptness in inherent informative descriptions for images (or visual descriptions). Despite the facilitation of domain transferability geared by extensive pretraining and fine-tuning on large-scale datasets, the integration of foundation models alongside this approach incurs a significant computational overhead. Recent lightweight captioning methods~\cite{ramos2023smallcap} tend to minimise the computational burden by selectively updating a limited number of parameters during the training process. The utilisation of pre-trained large multimodal models facilitates the enhancement of retrieval mechanisms in lightweight image captioning, demonstrating practical efficacy. These methodologies hinge upon a pretrained Contrastive Language-Image Pre-training (CLIP) model~\cite{radford2021learning} to retrieve captions from a dataset that closely resemble the input image. However, CLIP encounters the \emph{modality gap phenomenon}~\cite{liang2022mind}, characterised by the existence of inherent noise and discrepancies between image-text pairs in web-based datasets, thus impacting the efficacy of learning a common semantic space and affecting dense retrieval in a multimodal setting. Addressing this disparity necessitates existing image captioning approaches to undergo training in an end-to-end fashion~\cite{mokady2021clipcap,ramos2023smallcap,wang2023cropcap}. Our key objective is to effectively diminish the modality gap for the subsequent task of visual descriptions, while simultaneously maintaining competitive performance to alternative lightweight captioning techniques. Our approach leverages a linear mapping technique and utilises RAG to extract the most proximate textual descriptions corresponding to an image, to provide it together with a prompt as input to a large language model (LLM) for generating a fresh visual description.  Furthermore, we introduce a novel \emph{continuous refinement process} aimed at expanding the training dataset for our mapping technique by incorporating synthetic descriptions generated through our training-free approach. The notion of training-free is to only perform the mapping (instead of training certain layers of the neural network) from the visual embedding space to the textual space catered via a linear mapping applying a simple least squares solution~\cite{lu2013deep}. This approach offers a cost-effective and efficient means to improve retrieval-augmented captioning techniques employing a pre-trained LLM, thereby addressing the modality gap. Our training-free approach is assessed on two well-known multimodal datasets, MSCOCO~\cite{lin2014microsoft} and Flickr30k~\cite{young2014image}, yielding competitive results on both datasets while utilising only 1 million trainable parameters. We explore the generalisability of our approach across datasets, such as from MSCOCO to Flickr30k. Additionally, our continuous refinement process for thresholding generated descriptions marginally enhances performance, thereby corroborating the notion that textual descriptions can augment image captioning effectiveness. We discern that when optimising for the conventional reference-free metric~\cite{hessel2021clipscore}, our approach tends to produce hallucinated content. Conversely, when filtering based on reference-based measures, such as BLEU (B)~\cite{papineni2002bleu}, ROUGE (R)~\cite{lin2004rouge}, SPICE (S)~\cite{anderson2016spice} and CIDEr (C-D)~\cite{vedantam2015cider}, we notice consistent improvements across these performance measures.\\ 
Our main \textbf{contributions} therefore include: a) We propose an approach for lightweight training-free image content descriptions; b) we demonstrate that generated image descriptions, derived from pre-trained LLMs, can be utilised to enhance our methodology further in the subsequent task of image captioning; and c) we evaluate our linear mapping approach for retaining the semantics  of images in the textual embedding space alongside supervised measures (i.e.\ involving human references) via a user-behaviour-driven metric, such as normalised Discounted Cumulative Gain (nDCG).%~\cite{jarvelin2002cumulated}.
\begin{figure}
\includegraphics[width=\linewidth]{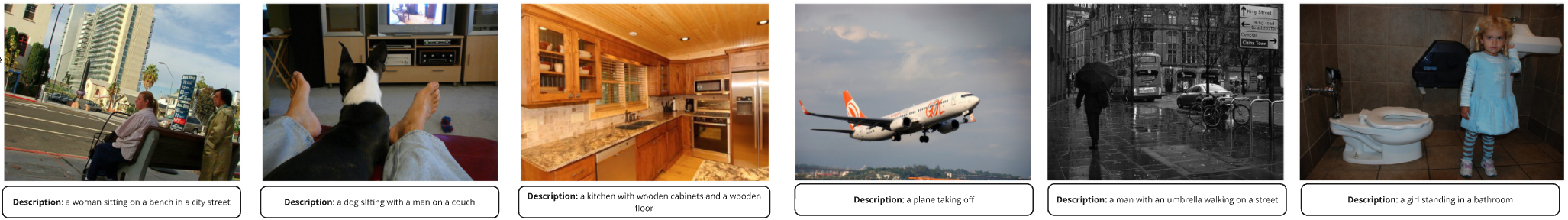}
\vspace{-20pt}
  \caption{Our method (\emph{mRAG-gim}) generates textual descriptions using MSCOCO validation images, employing our computed mapping derived from the MSCOCO training set.}
 \label{fig:gendes}
 \vspace{-20pt}
\end{figure}

\subsection{Prior Work}
\vspace{-4pt}
\textbf{Multimodal Semantics:} Combining multiple semantic spaces for an image feature representation can enhance retrieval performance. Recent advancements in LLMs have spurred investigation into conditioning them on visual input to improve their generation abilities, achieved via cross-attention mechanisms between pretrained unimodal models~\cite{li2022blip,alayrac2022flamingo,yang-etal-2023-vilm} or training a mapping network among the images and LLM input space~\cite{blip2,huang2023language,palme}. 
%This can be achieved through cross-attention mechanisms between pretrained unimodal models~\cite{li2022blip,alayrac2022flamingo,yang-etal-2023-vilm} or training a mapping network among the images and LLM input space~\cite{blip2,huang2023language,palme}. 
This led to the emphasis for pretraining on extensive datasets of paired image-text data, then fine-tuning for image captioning~\cite{tan2019lxmert,zhang2021vinvl}. Some approaches also use contrastive learning for semantic alignment between image and text modalities~\cite{radford2021learning}. Our method enhances retrieval-augmented text generation from images utilising ordinary least squares mapping. Unlike some approaches that rely solely on text data for image captioning~\cite{ijcai2023p481}, we ground images to textual descriptions via linear mapping addressing the modality gap. While other methods adjust pre-training objectives to align image and text modalities better~\cite{goel2022cyclip}, they were trained on smaller datasets than CLIP. Thus, we employ CLIP for retrieval and perform a linear mapping for grounding the visual feature space.\\
\textbf{LMMs for Visual Descriptions:} 
Image-language pre-training exemplified by CLIP~\cite{radford2021learning} garnered significant attention for its zero-shot learning and transfer capabilities. Recent advances, such as e-CLIP~\cite{shin2022clip} and DeCLIP~\cite{li2022supervision}, leverage self-supervision and multimodal supervision to enhance performance. However, large-scale contrastive pre-training~\cite{jia2021scaling,radford2021learning,schuhmann2022laion} demands datasets of hundreds of millions to billions of samples, yet the utilisation of noisy web text remains suboptimal for this purpose. SmallCap~\cite{ramos2023smallcap} introduced retrieval-augmentation utilising a pretrained LMM and cross-attention mechanism trained end-to-end. 
%LiT suggests that locking a well-pretrained image encoder preserves visual representations from noisy language supervision~\cite{zhai2022lit}. However, LiT's inability to align complex text with a fully-trained image encoder leads to suboptimal performance in cross-modal retrieval tasks. 
BLIP~\cite{li2022blip} employs a bootstrapped image-grounded text encoder to eliminate noisy captions, yet requires fine-tuning resulting in increased model parameters. Unlike previous methods, we explore the use of LLMs to align images with descriptions from the training dataset within the joint embedding space of CLIP~\cite{radford2021learning}, thereby mitigating the modality gap. This approach augments the retrieval mechanism, requiring only the provision of retrieved textual descriptions to the language model. All these approaches optimise the mapping from image space to the embedding or hidden space of the pretrained LLM.
%This approach serves to augment the retrieval mechanism, requiring only the provision of retrieved textual descriptions to the language model. All these approaches optimise the mapping from image space to the embedding or hidden space of the pretrained LLM.
\vspace{-8pt}
\section{Preliminaries}
We employ a source dataset $\mathcal{S}$ (MSCOCO), containing image-text pairs ($X_i , T_i$). Initially, we employ a CLIP image encoder (IE) $\Phi_{IE}:p\rightarrow \mathbb{R}^d$ that generates a visual embedding matrix $\mathbb{V}_{S_{train}} = (v_1,\ldots, v_n)^{T}\in \mathbb{R}^{n\times d}$, which embeds images from the training subset $S_{train}\subset \mathcal{S}$, where $p$ represents the pixel space and $d$ represents the dimension of the embedding space. Subsequently, we leverage the CLIP text encoder ($TE$) $\Phi_{TE}: D\rightarrow \mathbb{R}^{d}$ to embed the associated textual descriptions ($D$), resulting in $\mathbb{E}_{S_{train}} = (e_1,\ldots,e_n)^{T}\in \mathbb{R}^{n\times d}$, where $n$ signifies the number of embedded textual descriptions in $S_{train}$. Given varied textual descriptions accompany each image in MSCOCO~\cite{lin2014microsoft}, we consider each instance of the image to be a distinct occurrence within the source dataset~$\mathcal{S}$. Then, we employ a least squares linear model (Ordinary Least Square (OLS)) $L_{M}\in\mathcal{R}^{d\times d}$, utilising the features $v_i \in \mathbb{V}_{S_{train}}$ as an input and the corresponding textual embeddings $e_i \in \mathbb{E}_{S_{train}}$ as targets, with the objective of minimising $\left\|L_M v_i - e_i \right\|^2 \forall (X_{i},T_{i})\in{S_{train}}$. This linear mapping scheme serves to reconcile the modality gap existing among image and text semantics. The retrieval process has linear complexity ($O(n)$) and the computational cost of the linear mapping are cubical~$O(d^3)$, with $d$ representing the dimensionality of the LMM (CLIP) semantic space and $n$ denoting the quantity of embeddings stored in the vector database. Our approach is not constraint to this scenario as fitting a least square solution on a CPU can be done within minutes, since we pre-compute the embeddings for the vector database used in our pipeline. 
%Note that adopting an algorithmic viewpoint~\cite{chen2017online} can speed up the retrieval process. 

\vspace{-8pt}
\section{Our Approach: mRAG-gim} 
We propose a novel approach, namely multimodal Retrieval Augmentation for grounding images (\emph{mRAG-gim}), employing linear mapping to retain multimodal semantics within a unified embedding space. mRAG-gim utilises pretrained LMMs to retrieve relevant textual descriptions that correspond to a given input image. These descriptions are then passed to a pretrained LLM using a prompt engineering technique, wherein they serve as prompts for generating new visual descriptions. However, the effectiveness of retrieval is hindered by the recognised modality gap inherent in pretrained LMMs. mRAG-gim aims to bridge this gap through a lightweight linear mapping, computable via the least squares criterion~\cite{lu2013deep,mikolov2013distributed} for the closed-form solution. Therefore, this mapping can be tailored to the specific dataset (or task), thereby removing the need for end-to-end fine-tuning.
\begin{figure*}
\includegraphics[width=\linewidth]{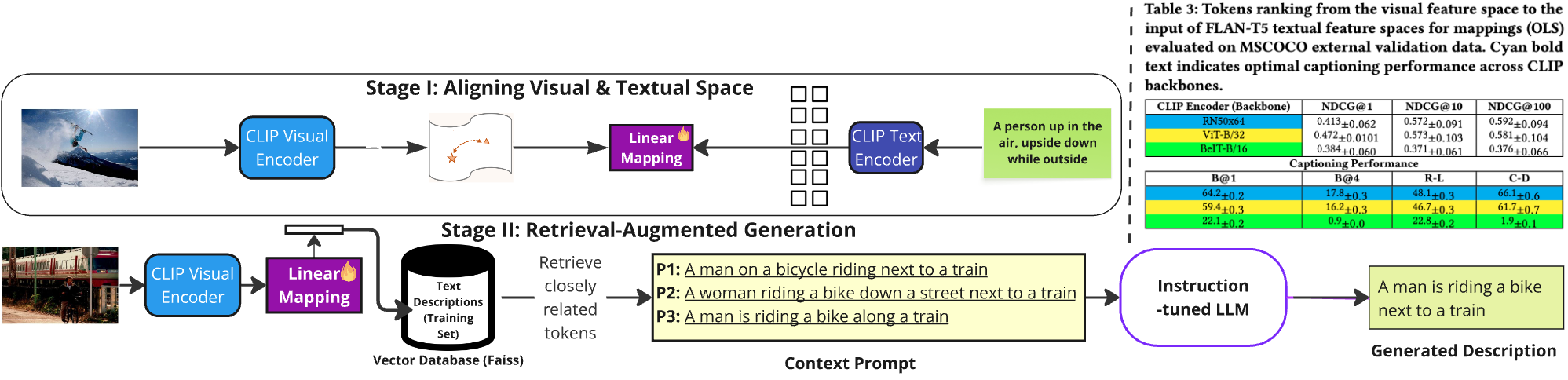}
\vspace{-16pt}
  \caption{The \emph{mRAG-gim} ground images to text by first establishing a linear mapping between visual and LLM token spaces, then using this mapping from stage I for inference to retrieve nearest-neighbour tokens that prompt FLAN-T5 for description generation.}
 \label{fig:lmap}
\end{figure*}
%The mRAG-gim pipeline for grounding images to textual descriptions follows a two-fold process. Initially, a linear mapping is performed via token encoding of the visual embedding space with the LLM space for the textual description counterparts. Then, we employ the estimated mapping from stage I for inference to retrieve nearest neighbouring tokens that closely matches the input image, which are fed as an input prompt to FLAN-T5 for generating descriptions.
%\subsection{Our Approach: mRAG-gim}
%\vspace{-10pt}
\begin{algorithm}%[hbt]
\caption{mRAG-gim: RAG-based Visual Descriptions}\label{alg:vistexts}
\footnotesize
\tiny
\begin{algorithmic}[1]
%\vspace{-3pt}
\Require Image encoder $\Phi_{IE}$, text encoder~$\Phi_{TE}$, training data~${S_{train}}$=$(X_i,T_i)$, test data~${S_{test}}=(X_j)$, LLM($\cdot$) as generative model, hyperparameter $k$, prompt~$\mathcal{P}$
\State $\left\{v_i,e_i \right\}_{i=1}^{\left|S_{train} \right|}\leftarrow\Phi_{IE}(X_i),\Phi_{TE}(T_i)$ for $(X_i, T_i)\in S_{train}$\Comment{Incorporate training data}
\State $L_M \leftarrow \text{fit linear mapping}(\left\{v_i,e_i \right\}$) \Comment{pre-compute mapping scheme}
\State $\mathbf{VD}\leftarrow\left\{e_i \right\}$ \Comment{Initialise vector database with training text descriptions}
\State $\left\{L_{M} v_i \right\}_{i=1}^{\left|S_{test}\right|}\leftarrow\Phi(X_j)$ for $(X_j, T_j)\in S_{test}$ \Comment{Incorporate unseen images from test set in CLIP space}
\State $\{\mathcal{D}_j\}\leftarrow\text{topk}(\{L_M v_j\},\mathbf{VD},k) $ \Comment{top-k descriptions to retrieve from vector database}
\State $\{\mathcal{G}_j\}\ \leftarrow\text{LLM}(\text{concatenate}(\mathcal{P}+\mathcal{D}_j))$ \Comment{Generate new textual descriptions}
\Ensure $\{\mathcal{G}_j\}$
\end{algorithmic}
%\vspace{-10pt}
\end{algorithm}
%\vspace{-10pt}
We schematically describe our approach in Fig.~\ref{fig:lmap}. The pipeline delineates two stages, where initially (Stage I) our semantic linear mapping ($L_M$) enables the coupling of an image encoder with a generative language model, facilitating the generation of language conditioned on image input. 
%Let \mathcal{D} represent the collection of tokens embedded with CLIP, acquired through the utilisation of the CLIP tokenizer on LMM. 
%We frame our task, where 
In Stage II, an input image $X\in p$ is followed by an embedding $v = \Phi_{IE}(X)$ and retrieving the top-$k$ textual descriptions within the set $\mathcal{D}$ via the cosine similarity ($\cos$), where $\mathcal{D} = \arg\max_{i\in 1,\ldots,n}^{k}\cos(e_i, L_{M}v)$.
%\vspace{-12pt}
%\begin{align}

%\end{align}%}
%\vspace{-2pt}
%where $\cos$ depicts the cosine similarity. 
The textual descriptions obtained within the set $\mathcal{D}$ are presented to an LLM (FLAN-T5)~\cite{chung2022scaling} as a context prompt, accompanied by a prompt instructing the LLM to produce a new description for the image~$X$. We describe our approach for generating visual descriptions through mRAG-gim in Algorithm~\ref{alg:vistexts}. We further extend this algorithm to facilitate continuous refinement for retrieval-augmentation by utilising generated textual descriptions. Algorithm~\ref{alg:crefine} exclusively follows the training algorithm and is not employed for inference or evaluation (we report the evaluation of employed LLMs in Table~\ref{tab:tab1}). Our reference set comprises solely pre-computed texts. Throughout the continuous refinement process, generated descriptions are embedded once and appended to the reference set. We can enhance the dataset \(S_{\text{train}}\) by incorporating generated textual descriptions for images already present in the training set. The goal is to selectively include high-quality generated descriptions in \(S_{\text{train}}\) to enhance the overall performance of our model's predictions. To achieve this, we rely on an image captioning measures~$\tau$ among BLEU@4, ROUGE-L, CIDEr-D, and SPICE, which evaluate a candidate description against a set of reference textual descriptions and outputs a single numerical score.
%\vspace{-10pt}
\begin{algorithm}%[H]
%\centering
\caption{Continuous Refinement for Retrieval Augmentation}\label{alg:crefine}
\footnotesize
\tiny
\begin{algorithmic}[1]
\Require Image encoder $\Phi_{IE}$, text encoder~$\Phi_{TE}$, training data~${S_{train}}$=$(X_i,T_i)$, validation data~${S_{val}}=(X_j)$, LLM($\cdot$) as generative model, hyperparameter $k$, prompt~$\mathcal{P}$
\State $\left\{v_i,e_i \right\}_{i=1}^{\left|S_{train} \right|}\leftarrow\Phi_{IE}(X_i),\Phi_{TE}(T_i)$ for $(X_i, T_i)\in S_{train}$  \Comment{Incorporate training data}
\State $L_M \leftarrow \text{fit linear mapping}(\left\{v_i,e_i \right\}$) \Comment{Pre-compute mapping scheme}
\State $\mathbf{VD}\leftarrow\left\{e_i \right\}$ \Comment{Initialise vector database with training text descriptions}
\State $\bar{\tau}\leftarrow\text{eval}(S_{val},\Phi_{IE},L_{M},\text{LLM},\mathbf{VD})$\Comment{Evaluation on the validation data}
\For {$ \_~ \text{in range}(n)$}
    \State $\{\mathcal{D}_i\}\leftarrow\text{topk}(\{L_M v_i\},\mathbf{VD},k)$\Comment{Top-k descriptions to retrieve from vector database}
    \State $\{\mathcal{G}_j\}\ \leftarrow\text{LLM}(\text{concatenate}(\mathcal{P}+\mathcal{D}_j))$ \Comment{Generate new textual descriptions}
    \State $\{\mathcal{G}_j\}\ \leftarrow\text{filter}(\mathcal{G}_j, \bar{\tau})$\Comment{Remove de-duplicate descriptions based on $\bar{\tau}$}
    \State $\left\{e_l \right\}_{l=1}^{\left|S_{train} \right|}\leftarrow\Phi_{IE}(u_l)$ for $(u_l)\in \mathcal{D}$\Comment{Incorporate new generated descriptions}
    \State $\mathbf{VD}\leftarrow\mathbf{VD}\cup\left\{e_l \right\}$\Comment{Add generated descriptions to database}
    \State $\left\{v_i,e_i,e_l \right\}_{i,l=1}^{\left|S_{train} \right|}\leftarrow e_{l}$ for $(e_{l})\in \left\{e_l \right\}$  \Comment{Augment training data}
    \State $L_M \leftarrow \text{fit linear mapping}(\left\{v_i,e_i,e_l \right\}$) \Comment{Recompute mapping scheme}
    \State $\bar{\tau}\leftarrow\text{eval}(S_{val},\Phi_{IE},L_{M},\text{LLM},\mathbf{VD})$\Comment{Update average metric}
\EndFor
%%\Ensure $\{\mathcal{G}_j\}$
\end{algorithmic}
\end{algorithm}
%\vspace{-10pt}
Initially, we assess mRAG-gim's performance on the validation set by calculating the average measure~$\bar{\tau}$, which gives us an indication of how well the generated descriptions match the expected quality. Following that, we produce a batch of new textual descriptions for images in ($S_{\text{train}}$) by randomly selecting from the LLM. For each generated description, we calculate its individual metric score using $\tau(\cdot,\cdot)$, and discard any textual descriptions that fall below the computed average score~$\bar{\tau}$. Once we generate synthetic textual descriptions for all images in the $S_{\text{train}}$ dataset, we include them in our training data and vector database, and then retrain $L_M$. Then, we assess the performance of the updated $L_M$ on the validation set and adjust the average performance metric~$\bar{\tau}$. We iterate through this procedure multiple times until we cease to observe any further enhancements in~$\bar{\tau}$.
\vspace{-12pt}
\section{Experiments}
We evaluate our approach on an image captioning task conducted on the widely known text-image datasets MSCOCO~\cite{lin2014microsoft} and \linebreak Flickr30k~\cite{young2014image}. 
%We investigate the transferability of our approach across domains, specifically from MSCOCO to Flickr30k. 
We conduct ablation analyses on our mapping technique, exploring various levels of linguistic supervision at the token level to determine the optimal configuration. We evaluate our linear mapping method by ranking tokens (shown in Table 3 that exhibit lexical correspondence with the reference descriptions of images.
%\vspace{-14pt}
\begin{table*}[h!]
%\small
\caption{\label{tab:res1} Quantitative comparison against state-of-the-art methods on two multimodal datasets. $\dagger$ depicts the training time on CPU, and $\dagger\dagger$ is the training time in multiple of n iterations to run the Algorithm~\ref{alg:crefine}. The metric scores on MSCOCO and Flickr30k test sets are in-domain and out-domain data ($\uparrow$ is better). The CLIP-score (C-S) and RefClip (RC)~\cite{hessel2021clipscore} is for the ablation analysis. Bold (in cyan) indicates the best results among baseline methods. The green shaded cells are for mRAG-gim without linear mapping, whereas the yellow shaded cells are with linear mapping.}%CIDEr-D (C-D), and SPICE (S)
\vspace{-10pt}
\begin{adjustbox}{width=\linewidth}
    \begin{tabular}{|c|ccccc|c|cc|ccc|}\cline{2-12}
    \hline
    \multirow{2}{*}{\textbf{\diagbox{Method}{Datasets}}} &
    \multicolumn{3}{c|}{\textbf{MSCOCO}} & 
    \multicolumn{2}{c|}{\textbf{Flickr30k}} & 
    \multicolumn{1}{c|}{\textbf{Training}} & 
    \multicolumn{2}{c|}{\textbf{Ablation  (CIDEr-D)}} & 
    \multicolumn{3}{c|}{\textbf{Ablation Test (CLIP-score/RefClip)}} \\
     & B@4 & S & C-D & S & C-D & (Params,Time) & in-domain & out-domain & C-D & S & C-S/RC\\
    \cline{2-12}
    \hline
    ClipCap~\cite{mokady2021clipcap} & 33.5 & 21.1 & 113.1 & 15.8 & 57.9 & 43M,1.4hr(L4) & ~-~ & ~-~ & \cellcolor{green}15.8$_{\pm0.1}$ & \cellcolor{green}5.8$_{\pm0.1}$ & \cellcolor{green}75.6, 79.1 \\
    SmallCap~\cite{ramos2023smallcap} & 36 & 21 & 117.4 & ~-~ & ~-~ & 1.8M,13hr(L4) & 55.4 & \cellcolor{cyan}52.2 & \cellcolor{green}17.8$_{\pm0.2}$ & \cellcolor{green}6.6$_{\pm0.2}$ & \cellcolor{green}75.9, 79.4  \\
    Llama-AdapterV2~\cite{gao2023llama} & 36.2 & ~-~ & 122.2 & ~-~  & ~-~ & 14M,~-~ & ~-~ & ~-~ & \cellcolor{green}80.0$_{\pm0.7}$ & \cellcolor{green}18.4$_{\pm0.1}$ & \cellcolor{green}79.3, 80.2 \\
    mRAG-gim & \cellcolor{cyan}31.3$_{\pm0.2}$ & \cellcolor{cyan}21.1$_{\pm0.2}$ & \cellcolor{cyan}107.8$_{\pm0.4}$ & \cellcolor{cyan}16.1$_{\pm0.3}$ & \cellcolor{cyan}64.5$_{\pm1.7}$ & 1M,8s-10s$^\dagger$ & 56.9$_{\pm1.4}$ & 43.2$_{\pm1.2}$ & \cellcolor{yellow}47.0$_{\pm0.3}$ & \cellcolor{yellow}14.1$_{\pm0.2}$ & \cellcolor{yellow}73.7, 78.1 \\
    mRAG-gim+Alg.~\ref{alg:crefine} & 29.0$_{\pm0.4}$ & 21.6$_{\pm0.1}$ & 104.3$_{\pm0.7}$ & \cellcolor{cyan}17.3$_{\pm0.1}$ & \cellcolor{cyan}66.8$_{\pm2.0}$ & 1M,8s-10s$^{\dagger\dagger}$ & \cellcolor{cyan}60.2$_{\pm1.4}$ & 45.1$_{\pm1.5}$ & \cellcolor{yellow}41.4$_{\pm0.2}$ & \cellcolor{yellow}12.9$_{\pm0.1}$ & \cellcolor{yellow}72.8, 77.2 \\ %\midrule
    \hline
    \end{tabular}
    \end{adjustbox}
\end{table*}
\vspace{-10pt}
\subsection{Implementation Details \& Results}
The datasets are partitioned into training, validation, and test sets following the standard protocol~\cite{karpathy2015deep}. Prior to further analysis, we conducted preprocessing steps, including length normalisation and mean centering of both image and textual embedding vectors in line with related strategies in~\cite{artetxe2016learning}. We found the mean centering of the embedding spaces to be of particular significance. Also, we computed the mapping on image-text pairs within the training subset using OLS. We observe that our mapping via OLS between CLIP backbone models and FLAN-T5 achieves the optimal alignment among visual and textual embedding spaces, and receives the highest NDCG score reported in Table 3. Moreover, the parameter count for this mapping is contingent upon the dimensionality, $d$ ($\leq$1024). To optimise image captioning performance, we conduct an exhaustive search across various image encoders, language models, decoding techniques, and prompting strategy. We adopt a vector database, Faiss~\cite{johnson2019billion} 
for storing data artefacts due to its capability for efficient handling of storage and retrieval tasks within vector databases. Our final configuration incorporates an RN50x64 CLIP encoder alongside a large instance of FLAN-T5 model~\cite{chung2022scaling}. We utilise FLAN-T5 for generating descriptions and employ a consistent prompting strategy akin to prior studies. Specifically, our prompt instruction reads as follows: \say{Show similar images: {} The image describes: }, with the insertion of the most analogous descriptions in place of the placeholder. We conduct experiments with prompt summarisation, albeit yielding slightly substandard results. In the context of Algorithm~\ref{alg:crefine}, we explore different metrics to assess the quality threshold of generated descriptions reported in Fig.~\ref{fig:gendes}. Our findings indicate that CIDEr-D~\cite{vedantam2015cider} is well-suited for this purpose and typically leads to a marginal enhancement across all other evaluation metrics. In Table~\ref{fig:lmap}, we present our approach results on MSCOCO. mRAG-gim with only 1 million trainable parameters, outperforms competitive baseline methods significantly in training time. Despite its lower training budget, mRAG-gim achieves performance close to SmallCap~\cite{ramos2023smallcap} in SPICE. However, there remains a significant gap in supervised (n-gram based) metrics. Inference time is similar for SmallCap and mRAG-gim (approximately 0.2 seconds on average on an L4 GPU). Employing our continuous refinement with mRAG-gim enhances SmallCap in terms of SPICE score, with a significant improvement observed after two iterations. mRAG-gim yields slightly lower scores in CIDEr-D and SPICE. Though, mRAG-gim with continuous refinement process effectively eliminates this discrepancy following one iteration. No further enhancements were observed thereafter. Conversely, there is a slight decline in SPICE after one iteration of Algorithm~\ref{alg:crefine}. We examine the transfer of mRAG-gim from MSCOCO to Flickr30k across two scenarios as part of an \emph{ablation test}, (a) transferring the mapping with in-domain data, and (b) transferring both embedding data artefacts and the mapping. We assess the impact of an orthogonality constraint on the mapping. Compared to SmallCap, the sole lightweight captioning method employing RAG, our approach using Algorithm~\ref{alg:crefine} achieves the highest CIDEr-D score surpassing SmallCap, while mRAG-gim shows only marginal improvement over SmallCap. This phenomenon is exclusive to transferring the mapping to untrained data, suggesting mRAG-gim's superior efficacy in utilising new data without requiring training.  
\vspace{-6pt}
\begin{table}[htb]
%\centering%Each model consists of eight layers, and the embedding size is 512. $\mid\theta\mid,\mid\theta_{v}\mid$  are the number of parameters.
%Evaluation of LLMs on the MSCOCO validation set provided standard error for all models.
\caption{LLM evaluations on the MSCOCO validation set provided standard errors for all models.}\label{tab:tab1}
\vspace{-8pt}
\footnotesize
\begin{adjustbox}{width=\linewidth}
\begin{tabular}{|c|c|c|c|c|c|c|}
\hline
\textbf{LLMs} &  \textbf{B@1} & \textbf{B@2} & \textbf{R-L} & \textbf{C-D} & \textbf{S} & \textbf{Params}\\
\hline
FLAN-T5-small & 57.6$_{\pm0.2}$ & 21.2$_{\pm0.2}$ & 54.2$_{\pm0.2}$ & 90.2$_{\pm0.6}$ & 20.6$_{\pm0.1}$ & 60M\\
FLAN-T5-base &  60.4$_{\pm0.2}$ & 22.5$_{\pm0.4}$ & 54.9$_{\pm0.2}$ & 92.3$_{\pm0.4}$ & 20.6$_{\pm0.1}$ & 220M\\
FLAN-T5-large & \cellcolor{cyan}77.7$_{\pm0.2}$ & \cellcolor{cyan}31.0$_{\pm0.5}$ & \cellcolor{cyan}57.9$_{\pm0.3}$ & \cellcolor{cyan}106.1$_{\pm0.7}$ & \cellcolor{cyan}21.0$_{\pm0.1}$ & 770M\\
FLAN-T5-xl & 76.0$_{\pm0.1}$ & 29.8$_{\pm0.3}$ & 57.0$_{\pm0.3}$ & 103.4$_{\pm1.0}$ & 20.6$_{\pm0.2}$ & 3B\\
FLAN-T5-xxl & 64.2$_{\pm0.1}$ & 23.5$_{\pm0.2}$ & 54.6$_{\pm0.1}$ & 94.7$_{\pm0.2}$ & - $_{\pm0.1}$ & 11B \\ \hline
DeBERTaV3-base~\cite{he2023debertav} & 59.2$_{\pm0.3}$ & 22.1$_{\pm0.4}$ & 54.7$_{\pm0.3}$ & 91.2$_{\pm0.2}$ & 20.9$_{\pm0.3}$ & 86M\\
LLaMa~\cite{touvron2023llama} & 63.8$_{\pm0.5}$ & 27.6$_{\pm0.5}$ & 50.2$_{\pm0.3}$ & 89.2$_{\pm0.9}$ & 19.4$_{\pm0.3}$ & 7B\\
\toprule
\multicolumn{7}{c}{\textbf{Results for the most similar-to-dissimilar (S2d) descriptions ordering in the Prompt}} \\
      \hline

Most S2d & 77.4$_{\pm0.1}$ & 30.4$_{\pm0.1}$ & 58.0$_{\pm0.3}$ & 105.6$_{\pm0.6}$ & 21.0$_{\pm0.3}$ &\\
d2S & 77.8$_{\pm0.1}$ & 30.5$_{\pm0.1}$ & 58.1$_{\pm0.3}$ & 106.3$_{\pm0.9}$ & 21.4$_{\pm0.3}$ &
\\
       \hline
\end{tabular}
\end{adjustbox}
\vspace{-12pt}
\end{table}
Our ablation analysis reveals that CLIP often assigns disproportionately high scores to low-quality captions generated by mRAG-gim tokens. We analyse Pearson correlations among evaluation metrics, expecting strong positive correlations between metrics and caption quality. However, while supervised metrics like C-D, R-L, and B@4 strongly correlate, CLIP-based metrics show only slight positive correlation. Surprisingly, SPICE is largely uncorrelated with other metrics in ablation test reported in Table~\ref{tab:res1}. Also, our finding in Table~\ref{tab:tab1} indicate that altering the order of descriptions in the prompt results in varying performances, thereby substantiating the presence of recency bias in large LLMs~\cite{zhao2021calibrate}. In the case of dissimilar-to-similar (d2S) ordering, the most similar descriptions initially emerge, resulting in a better performance in the CIDEr-D score.
%\vspace{-12pt}
%\begin{table}[htb]
%\centering
%\caption{Tokens ranking from the visual feature space to the input of FLAN-T5 textual feature spaces for mappings (OLS) evaluated on MSCOCO external validation data. Cyan bold text indicates optimal captioning performance across CLIP backbones.}\label{tab:tab2}
%\vspace{-10pt}
%\footnotesize
%\tiny
%\begin{adjustbox}{width=\linewidth}
%\begin{tabular}{|c|c|c|c|}
%\hline
%\textbf{CLIP Encoder (Backbone)} &  \textbf{NDCG@1} & %\textbf{NDCG@10} & \textbf{NDCG@100} \\
%\hline
%\cellcolor{cyan}RN50x64 & 0.413$_{\pm0.062}$ & 0.572$_{\pm0.091}$ & 0.592$_{\pm0.094}$ \\
%\cellcolor{yellow}ViT-B/32&  0.472$_{\pm0.0101}$ & 0.573$_{\pm0.103}$ & 0.581$_{\pm0.104}$ \\
%\cellcolor{green}BeIT-B/16&  0.384$_{\pm0.060}$ & 0.371$_{\pm0.061}$ & 0.376$_{\pm0.066}$ \\ \hline
%\hline

%5\multicolumn{4}{c}{\textbf{Captioning Performance}} \\ \hline

%\textbf{B@1} & \textbf{B@4}  & \textbf{R-L}  & \textbf{C-D} \\\hline
%\cellcolor{cyan}64.2$_{\pm0.2}$ & \cellcolor{cyan}17.8$_{\pm0.3}$ & \cellcolor{cyan}48.1$_{\pm0.3}$ & \cellcolor{cyan}66.1$_{\pm0.6}$ \\
%\cellcolor{yellow}59.4$_{\pm0.3}$ & \cellcolor{yellow}16.2$_{\pm0.3}$ & \cellcolor{yellow}46.7$_{\pm0.3}$ & \cellcolor{yellow}61.7$_{\pm0.7}$ & 
%\cellcolor{green}22.1$_{\pm0.2}$ & \cellcolor{green}0.9$_{\pm0.0}$ & \cellcolor{green}22.8$_{\pm0.2}$ & \cellcolor{green}1.9$_{\pm0.1}$ \\
%       \hline
%\end{tabular}
%\end{adjustbox}
%\end{table}
\vspace{-10pt}
\section{Conclusion}
We presented \emph{mRAG-gim}, a RAG-based approach that generates visual descriptions of images via a linear mapping to narrow the well-known modality gap. Given our mapping computed via least squares, we retrieve closest captions from the training set for a given image, which along with an instruction, serve as input to a generative language model for new captions. We show that our linear mapping retains the best ranked tokens which transfers well from the visual embedding space to textual space, performing better on the captioning task. Also, we propose a continuous refinement process to iteratively improve the mapping using captions generated by the LLM. We selectively retain high-scoring synthetic captions to augment the training set, enhancing captioning metrics. We demonstrate that unsupervised metrics like CLIP-score are susceptible to misleading cues in our ablation analysis. Our approach achieves competitive performance compared to existing lightweight image captioning methods. Importantly, our mapping enables the use of smaller LLMs (FLAN-T5)~\cite{fu2024tiny}, democratising image captioning for resource-constrained users.

\section*{GenAI Usage Disclosure}
We do not use any automated tools (LLMs or GenAI tool) in this paper from both writing and development perspective.
%\begin{acks}
%To Robert, for the bagels and explaining CMYK and color spaces.
%\end{acks}

\bibliographystyle{ACM-Reference-Format}
\bibliography{sample-base}

%\appendix

\end{document}